\pdfoutput=1

\documentclass[11pt]{article}

\usepackage{ifthen}
\newboolean{anonymous}   
\setboolean{anonymous}{false}   

\ifthenelse{\boolean{anonymous}}{\usepackage[review]{acl}}{\usepackage{acl}}
\usepackage{listings}
\usepackage{tablefootnote}
\usepackage{times}
\usepackage{latexsym}
\usepackage{amsmath}
\usepackage{booktabs}
\usepackage{caption}
\DeclareMathOperator*{\argmax}{arg\,max}

\usepackage{tikz}
\usetikzlibrary{arrows.meta}
\tikzset{%
  >={Latex[width=1mm,length=1mm]},
  base/.style = {rectangle, rounded corners, draw=black, text centered, font=\sffamily},
  transformer/.style = {base,  minimum width=5.5cm, minimum height=1.5cm, rotate=270},
  embedding/.style = {base, minimum width=0.5cm, minimum height=1cm},
  inputText/.style = {align=right, text width=2cm},
  glossText/.style = {align=left, text width=2cm},
  cc1/.style={circle,draw,node contents={\tiny \textbullet}},
  scoredot/.style={circle,draw,fill=gray!90},
}
\usepackage{hyperref}

\usepackage[T1]{fontenc}

\usepackage[utf8]{inputenc}
\usepackage{pythonhighlight}
\usepackage{microtype}
\usepackage{float}
\restylefloat{table}

\title{MWE as WSD: Solving Multiword Expression Identification with Word Sense Disambiguation}

\author{*Joshua Tanner \\
  Mantra Inc. \\
  \texttt{josh@mantra.co.jp} \\\And
  *Jacob Hoffman \\
  Project Ronin \\
  \texttt{jacob@projectronin.com} \\}

\begin{document}
\maketitle
\ifthenelse{\boolean{anonymous}}{}{{
\let\thefootnote\relax\footnote{{*Both authors contributed equally to this work.}}
\setcounter{footnote}{0}
}}

\begin{abstract}
Recent approaches to word sense disambiguation (WSD) utilize encodings of the sense gloss (definition), in addition to the input context, to improve performance. In this work we demonstrate that this approach can be adapted for use in multiword expression (MWE) identification by training models which use gloss and context information to filter MWE candidates produced by a rule-based extraction pipeline. Our approach substantially improves precision, outperforming the state-of-the-art in MWE identification on the DiMSUM dataset by up to 1.9 F1 points and achieving competitive results on the PARSEME 1.1 English dataset. Our models also retain most of their WSD performance, showing that a single model can be used for both tasks. Finally, building on similar approaches using Bi-encoders for WSD, we introduce a novel Poly-encoder architecture which improves MWE identification performance.
\end{abstract}

\section{Introduction}

Word sense disambiguation (WSD), the task of predicting the appropriate sense for a  word in context, and multiword expression (MWE) identification, the task of identifying MWEs in a body of text, both deal with determining the meaning of words in context \citep{maru-etal-2022-nibbling, constant-etal-2017-survey}. They have traditionally been treated as separate tasks, but this is potentially disadvantageous as WSD performed on words which are part of unrecognized MWEs cannot produce correct meanings, and the meanings of polysemous MWEs are ambiguous even after identification. For example, the sentence ``She inherited a fortune after her grandfather kicked the bucket'' tells us that someone's grandfather has died, but we would not expect to find meanings associated with death in the sense inventories of either \textit{kick} or \textit{bucket}. WSD  cannot capture the meanings of these words in context unless the relevant MWE is identified first. However, like many MWEs, \textit{kick the bucket} can have a literal, non-compositional meaning as in ``He kicked the bucket down the hill,'' so we also cannot indiscriminately mark all combinations of words in known MWEs as MWEs. MWEs can also have multiple possible senses in the same way words can:  \textit{break up} can refer both to objects physically breaking apart and romantic relationships ending, so even in cases where it is correctly identified as a MWE its meaning is ambiguous without WSD. Identifying the meanings of all words in a sentence requires solving these tasks together. 

WSD and MWE identification can be used in preprocessing to improve performance of downstream tasks such as machine translation or information extraction \cite{zaninello-birch-2020-multiword, song-etal-2021-improved-word, barba-etal-2021-esc}. They also have more direct applications in helping language learners -- for whom MWEs are particularly challenging  \cite{mweinlanguagelearning, mwehardforl2} -- understand the meaning of words or MWEs in context.

In this paper, we propose a system that tackles these tasks together, using a MWE lexicon and rule-based pipeline to identify MWE candidates and a trainable model to both perform WSD and filter MWE candidates. Our model is a modified Poly-encoder \citep{poly-encoders}, a natural extension of previous work using Bi-encoders for WSD \cite{blevins-zettlemoyer-2020-moving, kohli-biencoder}. Utilizing gloss information\footnote{For example, in ``The couple \textbf{broke up} amicably'', the gloss, or definition, of the sense of \textbf{break up} is ``discontinue an association or relation; go different ways.'' \cite{Miller1995}} allows our model to consider the meaning of MWEs and filter out candidates where the constituents of a MWE are present but the MWE meaning does not fit the context, such as the aforementioned literal usage of \textit{kick the bucket}. Our method improves precision and achieves state-of-the-art F1 for MWE identification on the DiMSUM dataset \citep{schneider-etal-2016-semeval} and competitive performance on the PARSEME 1.1 English data \citep{ramisch-etal-2018-edition}. To the best of our knowledge, this work is the first to use glosses as a resource for MWE identification. Our contributions are summarized as follows:

\begin{itemize}
\itemsep0em 
  \item We present a system which solves MWE identification and WSD together, achieving state-of-the-art results for MWE identification on DiMSUM and only 6\% less F1 for WSD than an equivalent single-task model
  \item We propose a novel Poly-encoder architecture which outperforms standard Poly-encoders on both tasks, and Bi-encoders on PARSEME MWE identification
  \item We explore why our system performs well and where it falls short through ablations and a detailed error analysis with examples 
\end{itemize}

\noindent We make all of our code, models and data public\ifthenelse{\boolean{anonymous}}{}{{\footnote{Code, data and links to models available at \href{https://github.com/Mindful/MWEasWSD}{https://github.com/Mindful/MWEasWSD}}}}.

\section{Related Work}

\subsection{Word Sense Disambiguation}

Until the last few years, most approaches to WSD treated senses only as one of many possible labels in a classification task. This formulation limits the information available to the model about each sense to only what is learnable from the training data, and can lead to poor performance on rare or unseen senses. To mitigate these problems, recent approaches have improved performance by incorporating sense glosses \citep{blevins-zettlemoyer-2020-moving, barba-etal-2021-esc, zhang-etal-2022-word}. 

Our work is inspired by this methodology and utilizes gloss information to improve MWE identification. In particular, \citet{blevins-zettlemoyer-2020-moving} demonstrate that a simple Bi-encoder model consisting of two BERT \citep{devlin-etal-2019-bert} models can achieve competitive WSD performance, with \citet{kohli-biencoder} improving Bi-encoder training for WSD and \citet{song-etal-2021-improved-word} achieving further performance gains through improved sense representations. Bi-encoder models are also particularly efficient at inference time because gloss representations can be computed in advance and cached.

\subsection{Poly-encoders}
The Poly-encoder architecture was proposed by \citet{poly-encoders} as a middle ground between Bi-encoders and Cross-encoders (which jointly encode all possible input pairs), retaining the speed advantage of the Bi-encoder, but allowing information to flow between the two encoder outputs like the Cross-encoder. It can be used in place of a Bi-encoder in tasks such as information retrieval \citep{li-etal-2022-using} text reranking \citep{kim-etal-2022-botstalk}, or in our case MWE identification and WSD.

\begin{figure*}
\includegraphics[width=16cm, height=10cm]{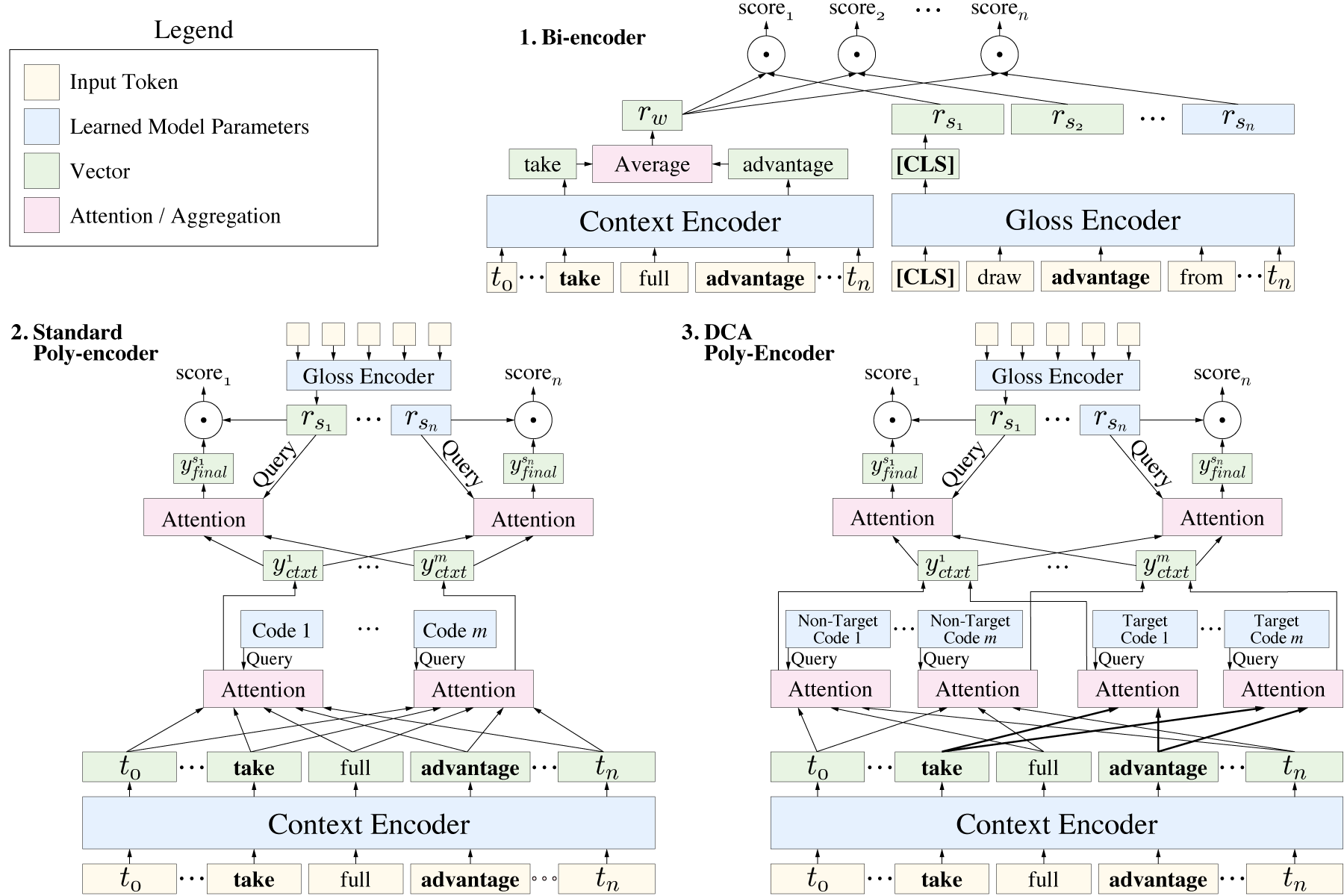}

  \caption{Each model scoring the MWE \textbf{take advantage}. ``Draw advantage from'' is the gloss for one possible sense. The gloss encoder produces sense representations $r_s$ using the [CLS] embedding in all models. The MWE representation $r_w$ is an average of constituents for the Bi-encoder and the combination of attention for each code for the Poly-encoder. The DCA Poly-encoder learns separate codes for target and non-target tokens, allowing it to attend differently to the MWE and surrounding context. Scores are the similarity between $r_s$ and $r_w$ computed as the dot product, and the model predicts the sense with the highest score. }
\label{figure:1}
\end{figure*}

\subsection{Multiword Expression Identification}

Precisely defining what constitutes a MWE has proven to be difficult \cite{maziarz-etal-2015-procedural}, but they can be broadly defined as groupings of words whose meaning is not entirely composed of the meanings of included words \citep{MWEPainInTheNeck, MWETextbook}. This includes idioms such as \textit{a taste of one's own medicine},  verb-particle constructions such as \textit{break up} or \textit{run down}, compound nouns such as \textit{bus stop}, and any other grouping of words with non-compositional semantics. In fact, a significant portion of noun MWEs are named entities \citep{savary-etal-2019-without}.

The task of MWE identification is locating these MWEs in a given body of text. Common approaches to solving MWE identification include rule-based systems \citep{foufi-etal-2017-parsing, pasquer-etal-2020-verbal}, CRF-based systems \cite{liu-etal-2021-lexical}, and token tagging systems \citep{rohanian-etal-2019-bridging}. Rule-based systems remain competitive with neural models in this task, and many systems including ours use MWE lexicons in order to identify MWEs, which \citet{savary-etal-2019-without} argue are critical to making progress in MWE identification. \citet{kurfali-ostling-2020-disambiguation} and \citet{kanclerz-piasecki-2022-deep} are similar to our work in that they frame the task of MWE identification as a classification problem, although neither use gloss information.

Among all the types of MWEs, verbal MWEs are particularly difficult to identify due to their surface variability --- constituents can be conjugated or separated so that they become discontinuous \citep{pasquer-etal-2020-verbal}. Much work on verbal MWE identification, especially in languages other than English, has been done as part of recent iterations of the PARSEME shared task \citep{ramisch-etal-2018-edition}.

\section{Methodology}
In this section, we explain how our models perform MWE identification and WSD, and how our MWE identification pipeline works.

\subsection{Bi-encoder}
Bi-encoders for WSD, as defined by \citet{blevins-zettlemoyer-2020-moving}, consist of two BERT \citep{devlin-etal-2019-bert} models: a \textbf{context encoder} $T_c$ and \textbf{gloss encoder} $T_g$, which embed the context and sense glosses into the same embedding space. Given an input sentence $c = (w_0,...w_n)$ containing the target words to disambiguate, we first tokenize it and use the \textbf{context encoder} to produce representations for each token. Because tokenization may break words or MWEs up into multiple subwords, word or MWE representations $r_w$ are computed as an average of all included subwords. 
\begin{gather*}
    T_c(c) = t_0, ...t_n \\
    r_{w} = \frac{1}{|w|}\sum_{t \in w}t 
\end{gather*}
Then, for each target word or MWE, the \textbf{gloss encoder} produces a sense representation $r_s$ for each possible sense by encoding its gloss and taking the [CLS] token embedding.
\[ \label{eq:sense_rep}
   r_s = T_g(g_s)[0]
\]
Scores corresponding to possible senses for each target word are computed as the dot product similarity of the word and sense representations, and the model predicts the highest scoring sense.
\begin{gather*}
    \phi(w, s_i) = r_w \cdot r_{s_i} \\
    pred(w) = \argmax_{s_i} \phi(w, s_i) : s_i \in S_w
\end{gather*} 
\subsection{Poly-encoder}
Like the Bi-encoder, the Poly-encoder has a \textbf{context encoder} $T_c$ for target word contexts and a \textbf{gloss encoder} $T_g$ for glosses. There is also a new set of parameters that \citet{poly-encoders} refer to as \textbf{code embeddings}, $Q$. These codes are used as queries to extract information from context representations produced by the \textbf{context encoder}. The inputs to the Poly-encoder are the same as to the Bi-encoder, sense representations $r_s$ are computed identically, and predictions are still the highest scoring sense. However, senses are scored differently. 

We take the last hidden state of the \textbf{context encoder} as the context representation $r_c = T_{c}(c)$, which we use along with the code embeddings $Q =  (q_{1}, ..., q_{m})$ in the first dot-product attention step (\textit{code context attention}) of the Poly-encoder. We use a different set of embeddings for single words and MWEs. The number of embeddings, $m$, is a hyperparameter and their dimensionality is the same as the encoders' hidden sizes. The context representation $r_c$ is used as both keys and values in this dot-product attention module, yielding a \textit{code attended context} $Y_{ctxt}$. The representation a code $q_i$ extracts is as follows:  
\begin{gather*}
    (w_0^{q_i}, ..., w_{n}^{q_i}) = softmax(q_{i}\cdot r_{c_1}, ... q_{i}\cdot r_{c_n})\\
    y_{ctxt}^{i} = \sum_{j=1}^{n} w_{j}^{q_i} r_{c_{j}}
\end{gather*}
Sense representations $r_{s}$ are then used as queries and the code-attended context representations $Y_{ctxt}$ are used as keys and values in a final dot-product attention module, yielding a \textit{gloss attended code-context}. For a given sense $s$ of a word or MWE:
\begin{gather*}
    (w_{1}, ..., w_{m}) = softmax(r_{s}\cdot y_{ctxt}^{1}, ..., r_{s}\cdot y_{ctxt}^{m}) \\
    y_{final}^{s} = \sum_{i=1}^{m} w_{i} y_{ctxt}^{i}
\end{gather*}
We then take the dot product of the gloss attended code-context $y_{final}$ and each gloss embedding $r_{s_0}, ...r_{s_k}$, yielding a score for each gloss: $\phi(w, s_i) =  y_{final} \cdot r_{s_i}$.

\subsection{Distinct Codes Attention}
Since the Poly-encoder was originally designed to compute \textit{sentence} representations, it contains no mechanism for explicitly focusing on a specific set of target words/subwords. To address this problem, we propose a variation of the Poly-encoder which we call ``distinct codes attention'' (DCA). We change the \textit{code context attention} step of the Poly-encoder so that it can attend differently to target words and the surrounding context, using two sets of code embeddings: one set for target words, $Q_t$ and one set for non-target words $Q_{nt}$. Since we also maintain different code embeddings for single words and MWEs, this gives us a total of four sets of code embeddings. 

In the first attention module, \textit{code-context attention}, we construct two key matrices, one to be used with the target code queries $Q_t$ and one to be used with the nontarget code queries $Q_{nt}$. First we create two masks which pick out target or nontarget subwords: the target mask $M_{t}$, which is $1$ at the indices of target subwords and $0$ otherwise, and the nontarget mask $M_{nt}$ which is the opposite. We then multiply each mask by the encoded context $r_{c}$ to get target and nontarget key matrices $K_{t} = M_{t}r_{c}$ and $K_{nt} = M_{nt}r_{c}$. Next we compute target and nontarget query results ($QK^T$) and add them.
\[
    QK^{T} = Q_{t}K_{t}^{T} + Q_{nt}K_{nt}^{T}
\]
Finally, we softmax and multiply $QK^{T}$ by the encoded context $r_c$ to yield the \textit{code attended context}, $Y_{ctxt} =  softmax(QK^T)(r_{c})$. The \textit{gloss attended code-context} and final scores are then computed identically to the standard Poly-encoder.

\subsection{MWE Identification Pipeline} \label{mwe_pipeline}
We use a rule-based pipeline inspired by \citet{kulkarni-finlayson-2011-jmwe} for MWE identification. First, we compute initial candidates as all combinations of words in a sentence whose lemmas correspond to a MWE in our lexicon. That is, any group of words that when lemmatized corresponds to a known MWE, regardless of order or location in the sentence, is a candidate. This ensures we rarely miss known MWEs, but also produces many false positives, such as: \textbf{in that} in ``\textbf{That} was back \textbf{in} 1954, 55 years ago.''

Next, we filter the candidates by removing MWEs which are out of order or too gappy ($>$3 words in between constituents), and optionally by discarding MWE candidates judged to be incorrect by our DCA Poly-encoder (or other) model. We refer to the combination of rule-based extraction and filters with no model as the \textit{rule-based pipeline}. Since the model is applied as a final filter after extraction and the other filters, it can only improve precision. 
While the heuristic filters involving order and gappyness exclude some valid MWEs as well, they empirically improved performance on development data, and the majority of exclusions made by these filters are correctly removing false positives from candidate generation as can be seen in Table~\ref{tab:filters}. Note that many candidates excluded by one filter would also be excluded by another filter later in the pipeline.

\begin{table}[ht]
    \centering
    \begin{tabular}{c c c c c}
        & \multicolumn{2}{c}{\small\textbf{PARSEME}}  & \multicolumn{2}{c}{\small\textbf{DiMSUM}} \\
        \toprule
         & TN & FN & TN & FN \\
         \midrule
         OrderedOnly & 1005 & 29 & 427 & 23 \\
         MaxGappiness & 1549 & 52 & 655 & 49 \\
    \end{tabular}
    \caption{Tokens excluded by rule-based filters. True negatives represent correct exclusions (I.E. false positive candidates), and false negatives incorrect exclusions.}
    \label{tab:filters}
\end{table}

\noindent In cases of overlap between candidates remaining after filtering, we use only the candidate judged to be most likely by our model (or least gappy, in pipelines without models).

\subsubsection{Model Filter}
Because all of our MWE candidates correpond to words (and consequently subwords) in the input sentence, we can produce a representation $r_{w}$ for each MWE candidate, along with scores for each of their possible senses, the same way we do for words. However, since no MWE has a sense corresponding to the case where that candidate is a false positive, we define a special sense $s_n$ representing the case where the candidate is not a MWE. Since $s_n$ has no gloss, we cannot use the \textbf{gloss encoder} to compute a representation for it, and instead make this representation a learnable parameter matrix $r_{s_n}$, with the same dimensionality as the model's hidden size. This can then be used in our model's scoring functions to compute a score for the candidate not being a MWE. When using a model to filter, we exclude any MWE candidates whose highest scoring sense is the ``not a MWE'' sense $s_n$, retaining only candidates for which the below is true:
\[
 \exists s_i \in S_w\  \phi(w, s_i) > \phi(w, s_n)
\]
Note that since this filtering process involves computing scores for all possible senses, it also effectively performs WSD on any polysemous MWEs.

\section{Experimental Setup}

\subsection{Lexicon}
We use WordNet \cite{Miller1995} as our MWE lexicon for all experiments, treating every entry including the character ``\_'' as a MWE. All sense glosses are taken from WordNet 3.0. 
\subsection{Training Data}\label{data}
We train our models on  SemCor \citep{miller-etal-1993-semantic}, a WSD dataset containing a total of 226,036 examples annotated with senses from WordNet. In order to make the data usable for MWE identification in addition to WSD, we preprocess it in the following ways. First, we explicitly mark any words whose lemma includes the character ``\_'' as MWEs such that during training the possible labels for these MWEs also include the ``not a MWE'' sense. Since some discontiguous MWEs in SemCor are labeled only on a subset of the included words, we add stranded constituents to their parent MWE by attaching nearby words whose lemmas match constituents missing from the labeled MWE\footnote{For example, in ``Are they encouraged to take full legal advantage of these benefits?'' (ID d000.s015), the verb \textit{take} is correctly labeled as the MWE \textit{take\_advantage}, but \textit{advantage} is not labeled as being part of any MWE, so we attach it.}.
Finally, because SemCor contains no labeled negative examples of MWEs --- instances where the constituent words of a MWE are all present but their meaning in context does not match any of the MWE senses --- we add these ourselves. We generate synthetic negative examples using the rule-based pipeline with its filters inverted to mark combinations of words whose lemmas correspond to a known MWE but are out of order or very gappy as negative examples whose gold label is the ``not a MWE'' sense. We randomly add negative examples in this fashion until they account for just over 50\% of the MWE examples in the training data.

To mitigate the risk of the model learning only the heuristics used to generate these synthetic negatives, we also manually annotate a small number of examples. We do this by running the rule-based pipeline (Section~\ref{mwe_pipeline}) on the SemCor data and annotating output MWEs with their appropriate sense from WordNet or the ``not a MWE'', sense based on context. Because we exclude words already marked as MWEs and many MWEs in SemCor have already been annotated, $>$50\% of the newly annotated examples are negative. 

\begin{table}[H]
    \centering
    \begin{tabular}{c c c}
         & Pos MWE & Neg MWE   \\
        \toprule
         SemCor & 12409 & 0 \\
         +Annotation & 12907 & 658 \\
         +\footnotesize Synthetic Negatives & 12907 & 14688 \\
    \end{tabular}
    \caption{SemCor after each addition of data}
    \label{tab:semcor_table}
\end{table}

\begin{table}[H]
\footnotesize
\centering\setlength\tabcolsep{2.5pt}
\begin{tabular}{p{3cm} c p{1.3cm}}
Context & MWE & Type \\
\toprule
What effort \textbf{do} you \textbf{make} to assess ... & \textbf{make do} & Synthetic Negative\\
..your \textbf{in} plant feeding \textbf{operation}? & \textbf{in operation} & Annotated Negative \\
...\textbf{works} full-time \textbf{on} some other assignment? & \textbf{work on} & Annotated Positive
\end{tabular}
\caption{Examples of each annotation type}
\end{table}

\subsubsection{Fine-tuning Data}
After training on SemCor, we fine-tune on the MWE identification data in DiMSUM/PARSEME. We use any labeled examples of MWEs which are in our lexicon as positive examples, and then run our full pipeline (rules+model filter) on the data and take incorrect outputs as negative examples. This means that all the negative training examples used in fine-tuning are false positives from the model itself, allowing the model to learn from its mistakes. Because PARSEME and DiMSUM are not annotated with sense information (only a binary labeling of MWE or not), we use the first sense from WordNet as the gold label for positive examples when fine-tuning. For both datasets, we use 10\% of the training data as our development set.

\subsection{Training}

Like \citet{blevins-zettlemoyer-2020-moving}, we train with cross-entropy loss. The difference is that for MWEs, there is one additional possible label representing the ``not a MWE'' case. Given a word or MWE $w$, its gold sense $g_s$, and $|S_w|=j$ possible senses in the lexicon, this formalizes to:
\begin{gather*}
    \mathcal{L}(w, g_s) = -\phi(w, g_s) + \mathrm{log} \sum_{x \in X} \mathrm{exp(} \phi(w, x) \mathrm{)} \\
        X = 
    \begin{cases}
        \{s_0,...s_j, n\} & \text{if MWE}\\
        \{s_0,...s_j\} & \text{otherwise}
    \end{cases}
\end{gather*}
We train for 15 epochs on SemCor and three epochs for fine-tuning, computing F1 on the WSD and MWE identification dev sets once per epoch and use the best performing model as our final model. Batch size and other hyperparameters such as learning rate were determined by hyperparameter search. Further implementation and training details can be found in Appendix~\hyperref[appendix:imp]{A}.

\subsection{MWE Identification Evaluation}

We evaluate our system on the English section of the PARSEME 1.1 Shared Task \citep{ramisch-etal-2018-edition} and the DiMSUM dataset \citep{schneider-etal-2016-semeval}. We do not evaluate on STREUSLE \citep{schneider-etal-2018-comprehensive} as it requires predicting lexical categories and supersenses\footnote{The STREUSLE evaluation script rejects input without appropriate lexical categories/supersenses}, while our system predicts only the presence or absence of MWEs. To measure WSD performance, we use the evaluation framework established by \citet{raganato-etal-2017-word} and evaluate on the English all-words task.

\subsubsection{PARSEME 1.1}
The PARSEME data focuses on verbal MWEs, containing 3471 sentences in the training set and 3965 in test. Because the data contains only verbal MWEs, when evaluating on PARSEME we limit the output of our pipeline to verbal MWEs. 

\subsubsection{DiMSUM}\label{dimsum_eval}
The DiMSUM data consists of online reviews, tweets and TED Talks which have been annotated with MWEs and other information. There are 4799 sentences in the training set, and 1000 in the test set. Because noun phrases are marked as MWEs in DiMSUM, when evaluating on DiMSUM our pipeline also marks consecutive nouns as MWEs. 

\subsection{WSD Evaluation}
Following standard practice, we use the SemEval-2007 dataset \citep{pradhan-etal-2007-semeval} as our dev set, holding out the remaining Senseval-02, Senseval-03, SemEval-2013, and SemEval-2015, as test sets \citep{palmer-etal-2001-english, snyder-palmer-2004-english, navigli-etal-2013-semeval, moro-navigli-2015-semeval}. 

\begin{table*}[t]
    \centering\setlength\tabcolsep{2.5pt}
    \begin{tabular}{ccc|ccc lr ccc}
        \multicolumn{6}{c}{PARSEME 1.1} & &
        & \multicolumn{3}{c}{DiMSUM} \\
        \cmidrule{1-6}\cmidrule{7-11}
        \multicolumn{3}{c|}{\textbf{MWE-based}}&\multicolumn{3}{c}{\textbf{Token-based}}&
        \multicolumn{2}{c}{\textbf{System}} &\multicolumn{3}{c}{\textbf{MWEs}}\\
        \textbf{P}&\textbf{R}&\textbf{F}&\textbf{P}&\textbf{R}&\textbf{F}&\multicolumn{2}{c}{ } &\textbf{P}&\textbf{R}&\textbf{F}\\
        \hline
        --&--&36.0&--&--&40.2 &  Taslimipoor+ \citeyearpar{taslimipoor-etal-2019-cross} & Kirilin+ \citeyearpar{kirilin-etal-2016-icl}&73.5&48.4&58.4 \\
        -- & -- & \textbf{41.9} & -- & -- & -- & Rohanian+ \citeyearpar{rohanian-etal-2019-bridging} & \citet{williams-2017-boundary} & 65.4 & \textbf{56.0} & 60.4  \\
        \hline        36.1&\textbf{45.5}&40.3&40.2&\textbf{52.0}&\textbf{45.4} &
        \multicolumn{2}{c}{Liu+ \citeyearpar{liu-etal-2021-lexical}}&47.9&52.2&50.0\\
        16.3 & 39.9 & 23.1 & 19.2 & 43.9 & 26.7 &
        \multicolumn{2}{c}{Rule-based Pipeline}&57.7 & 55.5 & 56.6\\
        28.2 & 38.5 & 32.5\small$\pm$0.4 & 30.7 & 39.0 & 34.3\small$\pm$0.4 &
        \multicolumn{2}{c}{Rules + DCA ($S$)}&70.9 & 53.0 & 60.6\small$\pm$0.1\\
         35.7 & 39.3 & 37.4\small$\pm$0.6 & 37.7 & 38.6 & 38.1\small$\pm$0.4 &
        \multicolumn{2}{c}{Rules + DCA ($S/D$)}&78.2 & 51.8 & \textbf{62.3}\small$\pm$0.1\\
        \textbf{47.1} & 33.8 & 39.4\small$\pm$0.3 & \textbf{48.3} & 32.1 & 38.6\small$\pm$0.2 & 
        \multicolumn{2}{c}{Rules + DCA ($S/P$)}&75.7 & 49.4 & 59.8\small$\pm$0.1\\
        45.4 & 33.2 & 38.3\small$\pm$0.1 & 46.9 & 31.9 & 38.0\small$\pm$0.2&
        \multicolumn{2}{c}{Rules + DCA ($S/P/D$)}&\textbf{80.4} & 49.5 & 61.3\small$\pm$0.4 \\
    \end{tabular}
    \caption{Test set results on PARSEME 1.1 English and DiMSUM for MWE identification. All DCA poly-encoder models function as a final filter after the rule-based pipeline. Training data is listed in parenthesis: $S$=SemCor, $P$=PARSEME, $D$=DiMSUM. For trainable models we report the mean ($\pm$ standard deviation for the F1 score) of three runs with random seeds. Because our system uses gold POS tags/lemmas to look up sense glosses, we compare against systems using gold information where available, such as for \citet{liu-etal-2021-lexical} and  \citet{kirilin-etal-2016-icl}.}
    \label{results_table}
\end{table*}

\section{Results}
Table \ref{results_table} shows MWE identification performance for the rule-based pipeline (Section~\ref{mwe_pipeline}), and the same pipeline with the DCA Poly-encoder included as a final filter for various training data. Comparisons to the Bi-encoder and standard Poly-encoder can be found in Section~\ref{mwe_ablations}, or in detail in Appendix~\ref{appendix:results}.

Our system achieves moderate performance on PARSEME and competitive performance on the DiMSUM trained only on the modified SemCor data. When fine-tuned on both MWE identification datasets it further improves, reaching state-of-the-art performance on DiMSUM. Systems fine-tuned on either PARSEME or DiMSUM alone perform even better on their corresponding test set, but worse on the other test set, likely due to differences in domain and MWE type between the datasets. 

High precision stands out as a strength of our approach, but it suffers from low recall --- even the rule-based pipeline with no model filter lags behind other systems in recall. We attribute this mainly to the issue of lexicon dependence described in Section~\ref{limit_section}; MWEs missing from our lexicon account for a majority of our false negatives as we show in our error analysis (Section~\ref{error_analysis_section}). These findings echo \citet{savary-etal-2019-without} on the importance of lexicons for MWE identification, and suggest that there is room to improve performance by expanding the lexicon. While it is difficult to pinpoint exactly why we achieve state-of-the-art F1 on DiMSUM and not PARSEME, one significant difference is that more than 40\% of the DiMSUM test set MWEs are noun phrases, most of which we can detect without relying on a lexicon (as described in Section~\ref{dimsum_eval}). For PARSEME, we must always rely on our lexicon.

\subsection{WSD Performance}
We compare performance on the English WSD all-words task to \citet{blevins-zettlemoyer-2020-moving}, a similar Bi-encoder system trained only for WSD. Recent work in WSD has achieved higher scores \citep{barba-etal-2021-consec}, but our goal is to understand how the addition of the MWE identification task affects WSD performance. 
\begin{table}[h]
    \centering\setlength\tabcolsep{2pt}
    \begin{tabular}{l c|c c}
    \textbf{System} & \textbf{F1} & \textbf{System} & \textbf{F1} \\
    \hline
    Blevins+ & 79.0 & PolyEnc ($S$) & 73.8\small$\pm$0.2 \\
    DCA ($S$) & 77.2\small$\pm$0.1 & BiEnc ($S$) & 77.4\small$\pm$0.6 \\
    DCA {\small($S/P/D$)} & 74.4\small$\pm$0.6 & BiEnc {\small($S/P/D$)} & 74.2\small\small$\pm$1.0 \\
    \end{tabular}
    \caption{English WSD all-words task F1. }
    \label{tab:wsd}
\end{table}

\noindent  Our system retains most but not all of its WSD performance: F1 is 2\% lower when trained on our modified SemCor data and 6\% lower when fine-tuned on PARSEME+DiMSUM. We attribute this drop in F1 from fine-tuning to potentially confusing labels in the fine-tuning data: the gold label of positive examples is always the MWE's first sense, which may be incorrect for polysemous MWEs, and as we show in Section~\ref{error_analysis_section}, many negative example MWEs actually have senses appropriate for the context they are in. Consequently, the model cannot rely entirely on matching sense glosses to input contexts for this data and may forget some knowledge useful for WSD. 

Comparing models, the DCA Poly-encoder outperforms the standard Poly-encoder on WSD, but its performance does not significantly differ from the Bi-encoder. We leave Poly-encoder architectures better suited for WSD to future work. 

\begin{table*}[htbp]
\footnotesize
\centering\setlength\tabcolsep{2.5pt}
    \begin{tabular}{c c p{2.5in} p{2.5in}}
          Dataset & Type & Sentence & Note \\
        \toprule
         PARSEME & FP & \textit{...were \textbf{propped up} on a foot-warmer, ...} & \textbf{prop up} never marked as MWE in dataset \\
        PARSEME & FN & \textit{\textbf{Never mind}, Mrs. Bray will join you later.} & \textbf{never mind} missing from lexicon \\
        PARSEME & FP & \textit{...his mind \textbf{drifted off} to the accounts...} &  \textbf{drift off} sense ``fall asleep'' does not apply \\
        PARSEME & TN & textit{...as we \textbf{sat} side \textbf{by} side...} & \textbf{sit by} sense ``be inactive'' does not apply  \\
        DiMSUM & FP & \textit{Aww, \textbf{thank you}.} & \textbf{thank you} marked as MWE in 4 other sentences \\
        DiMSUM & FN & \textit{All our dreams can \textbf{come true},...} & \textbf{come true} missing from lexicon \\
        DiMSUM & FN & \textit{...this was a \textbf{breathe of fresh air}.} &  Present in lexicon; model filter false negative \\
        DiMSUM & TN & \textit{...impact my wardrobe \textbf{has on} the environment.} & \textbf{have on} sense ``dress in`` does not apply \\
    \end{tabular}
    \caption{Representative errors (FP/FN) and incorrect MWEs successfully excluded by the model filter (TN) }
    \label{tab:error_examples}
\end{table*}

\section{Analysis}
\subsection{MWE Identification Ablations}
\label{mwe_ablations}
 
\begin{table}[H]
    \centering\setlength\tabcolsep{1.5pt}
    \begin{tabular}{l c c c c}
    \textbf{System} & \small\textbf{PARSEME} & $\Delta$ & \small\textbf{DiMSUM} & $\Delta$ \\
    \toprule
    Rules+DCA & 38.3 & -- & 61.3 & -- \\
    \small{-SemCor Data} & 26.0 & -12.3 & 56.8 & -4.5 \\
    {\small-Rule Filters} & 35.5 & -2.8 & 61.8 & +0.5 \\
    Rules+BiEnc & 36.5 & -1.8 & 61.3 & -- \\
    Rules+PolyEnc & 34.0 & -4.3 & 60.3 & -1 \\
    \midrule
    Rules & 23.1 & -- & 56.6 & -- \\
    \small{-Filters} & 14.4 & -8.7 & 47.7\tablefootnote{Output for the rule-based pipeline with no filters was invalid according to the DiMSUM grader and had to be approximated, so it may be off by 1-2 F1 points.} & -8.9 \\
    \end{tabular}
    \caption{MWE identification F1 for ablations. Aside from the ablation removing SemCor data, all models are trained on SemCor+PARSEME+DiMSUM. }
\end{table}
\noindent Pretraining using the modified SemCor data is important; training only on the MWE identification datasets substantially reduces performance. Intuitively, this can be thought of as the model needing to learn how to encode context words and sense glosses before learning to apply that knowledge to MWE identification. 

We also see that while removing rule-based filters from the DCA pipeline lowers PARSEME F1, it slightly raises DiMSUM F1, suggesting that the necessity of these filters depends on the data. However, removing the rule-based filters only works because the DCA Poly-encoder can accurately exclude false positives: removing the same filters from the purely rule-based pipeline results in a very low F1. Finally, the DCA Poly-encoder substantially outperforms the standard Poly-encoder (PolyEnc) on both datasets and surpasses the Bi-encoder on PARSEME, demonstrating that our DCA Poly-encoder model can improve MWE identification performance. 

\subsection{Error Analysis}\label{error_analysis_section}
We perform an error analysis on the output of our SemCor trained and fine-tuned models on both test sets, taking 50 false positives and 50 false negatives from each combination of model and dataset (for a total of 400 examples). Select examples can be seen in Table~\ref{tab:error_examples}, and detailed statistics about the outcome of our analysis can be found in Appendix~\hyperref[appendix:error]{B}. 

We find that for $>$80\%\footnote{Computed excluding false-positives from the DiMSUM noun phrase detector, which does not use the lexicon} of false positives a sense from our lexicon was appropriate for the given context, but the target words were not marked as a MWE in the data. Many of these MWEs were present in our lexicon but nowhere in the test set, suggesting discrepancies between the scope of what WordNet and these datasets respectively define as MWEs. However, there were also a number of false positives that \textit{are} marked as MWEs in other places in the dataset. This could happen if these combinations of words were only marked as MWEs when they had specific meanings or particularly non-compositional semantics, but this was not the case for the examples we examined. These results speak to the difficult and potentially subjective nature of annotating MWEs, and we hope to see work exploring this area in the future.

For false negatives, $>$85\% were cases where the target MWE was missing from the lexicon, confirming that the bottleneck for recall is our system's lexicon. For the majority of the remaining false negatives, an appropriate sense for the given context was present in our lexicon, meaning that these were failures of our MWE identification system and not the lexicon. However, the fact that errors in matching meaning to context account for $<$20\% of false positives and $<$15\% of false negatives shows that our model has successfully learned how to judge whether a group of words constitutes a MWE with a given meaning.  See Table~\ref{tab:error_examples} true negatives for examples of MWEs excluded based on meaning.

\section{Conclusion}
In this work, we present an approach to MWE identification using rule-based candidate extraction with a model filter, achieving strong results on the PARSEME 1.1 English data and state-of-the-art results for MWE identification on the DiMSUM dataset. Our system performs both MWE identification and WSD with the same model, demonstrating that these tasks can be tackled together. We also introduce a modified Poly-encoder architecture better suited to MWE identification.

Our system's strength is its high precision for MWE identification. We show its low recall to be a function of lexicon size, and in future work we intend to expand the lexicon by mining MWEs and generating glosses for them, which has the potential to substantially increase recall for lexicon-based systems. Improved approaches for multitask training of MWE identification/WSD models could also be valuable; the ideal pipeline would be competitive with state-of-the-art systems in both tasks, and not just MWE identification.

Ideal applications of our system include MWE identification when a lexicon of target MWEs is available, or cases where quickly performing both MWE identification and WSD is valuable, such as in language learning and assisted reading tools.

\section{Limitations}\label{limit_section}
While our system performs well, the output of our MWE pipeline is limited to MWEs that are present in our lexicon or detectable with simple rules. Furthermore, because our model uses gloss text as input, we cannot effectively filter MWE candidates without sense glosses. Consequently, our approach to MWE identification depends on the presence of a high-quality lexicon which includes MWE lemmas and sense glosses, making it ill-suited for scenarios where data like this may not be available yet, such as in low resource languages. However, we are optimistic that work in MWE discovery \citep{ramisch-etal-2010-mwetoolkit} and gloss/definition generation \citep{bevilacqua-etal-2020-generationary} will help to mitigate this problem by automating parts of the data creation process.

\ifthenelse{\boolean{anonymous}}{}{
\section{Acknowledgements}
The authors thank the KERNEL organization (DEEPCORE Inc.) for providing the GPUs that made this work possible, Shane Steinert-Threlkeld and Shonosuke Ishiwatari for feedback on early versions of the paper, and Chika Ohe for helping make Figure~\ref{figure:1}.
}

\bibliography{anthology,custom}
\bibliographystyle{acl_natbib}

 \appendix
 \label{appendix:imp}
 \section{Implementation Details}
Bi-encoder and Poly-encoder models are implemented and trained with Pytorch Lightning \citep{pytorch-lightning}, using pretrained BERT models from the Transformers library \citep{wolf-etal-2020-transformers}. In particular, we use \textit{bert-base-uncased} as the base model for both encoders. We define batch size by the number of training examples (words or MWEs to be labeled) in each batch, and keep this number constant by adjusting the number of sentences and/or masking out examples to save them for the next batch. Our effective batch size is 32. All models were trained on a single GeForce GTX TITAN X GPU, with hyperparameters tuned using Weights \& Biases \cite{wandb} to run random sweeps and track performance. Separate sweeps were run for the Bi-encider and Poly-encoder, each having a maximum of 20 runs and using early stopping to terminate runs with poor performance. Our total compute time was approximately 160 days, though this would have been significantly lower using a newer model of GPU. Our models have 220M parameters, and fully training for 15 epochs on the modified SemCor data takes approximately $1.2 \times 10^{17}$ FLOPS. We used Prodigy \cite{prodigy_montani_honnibal} as our annotation tool. Further detail, including all training hyperparameters and instructions for reproduction, can be found in our published code. 

 \label{appendix:error}
 \section{Error Analysis Details}
 This appendix contain details about the frequency with which we found various types of false positives or false negatives in our error analysis. 
\subsection{PARSEME}
In the table below, \textbf{Def?} represents the \% of false positives where a sense appropriate for the predicted MWE was present in our lexicon. \textbf{MWE?} represents the \% of false positives where the MWE was present in other sentences in the dataset, and the \% of false negatives where it was present in our lexicon, respectively. 
\begin{table}[H]
    \centering
    \begin{tabular}{c|c|c|c}
          & \multicolumn{2}{|c|}{False Positives} & False Negatives \\
          \hline
         \textbf{Model} & \textbf{Def?} & \textbf{MWE?}  & \textbf{MWE?}  \\
         SemCor & 90\% & 16\% & 6\% \\
         fine-tuned & 90\% & 34\% & 16\%  \\
    \end{tabular}
    \caption{PARSEME Error Analysis}
    \label{tab:parseme_error}
\end{table}

\subsection{DiMSUM}

Our results on DiMSUM are similar to those of PARSEME, except that for the system using the SemCor model 22\% of the false positives were from the rule-based consecutive noun tagger, with that number increasing to 56\% for the fine-tuned model (the false positive rate drops substantially after fine-tuning the filtering model as can be seen in Table~\ref{results_table}, which leads to these errors accounting for a higher percentage of total false positives). The \textbf{Def?} and \textbf{MWE?} percentages for false positives in the below table are computed excluding consecutive noun tagger false positives. 

\begin{table}[H]
    \centering
    \begin{tabular}{c|c|c|c}
          & \multicolumn{2}{|c|}{False Positives} & False Negatives \\
          \hline
         \textbf{Model} & \textbf{Def?} & \textbf{MWE?}  & \textbf{MWE?}  \\
         SemCor & 92\% & 56\% & 4\% \\
         fine-tuned & 81\% & 63\% & 12\%  \\
    \end{tabular}
    \caption{DiMSUM Error Analysis}
    \label{tab:DiMSUM_error}
\end{table} 

\section{Detailed Performance}
Table~\ref{tab:appendix} below contains full scores for systems omitted from the main paper for brevity. 
\label{appendix:results}
{\setlength{\tabcolsep}{0.25em}\footnotesize
\begin{table*}[h] \centering
\begin{tabular}{p{3cm} | c c c | c c c | c c c | c}
System & \multicolumn{6}{|c|}{PARSEME 1.1} & \multicolumn{3}{|c|}{DiMSUM} & WSD \\
 \hline
  & \multicolumn{3}{c|}{\textbf{MWE-Based}} & \multicolumn{3}{c|}{\textbf{Token-based}} & \multicolumn{3}{c|}{\textbf{MWEs}} & \\
 & \textbf{P} & \textbf{R} & \textbf{F1} & \textbf{P} & \textbf{R} & \textbf{F1} & \textbf{P} & \textbf{R} & \textbf{F1} & \textbf{F1}\\
Rules (no filters) & 8.8 & 38.9 & 14.4 & 12.0 & 49.9 & 19.3 & 40.5$^{*}$ & 58.0$^{*}$ & 47.7$^{*}$ & -- \\
Rules (both filters) & 16.3 & 39.9 & 23.1 & 19.2 & 43.9 & 26.7 & 57.7 & 55.5 & 56.6 & -\\
BiEnc ($S$) & 27.5 & 38.8 & 32.2\small$\pm$0.8 & 30.0 & 39.43 & 34.1\small$\pm$0.3 & 70.7 & 52.57 & 60.0\small$\pm$0.4 & 77.4\small$\pm$0.6 \\
BiEnc ($S/P/D$) & 44.5 & 31.0 & 36.5\small$\pm$0.9 & 46.4 & 30.5 & 36.2\small$\pm$0.8 & 80.9 & 49.3 & 61.3\small$\pm$0.4 & 74.2\small$\pm$1.0 \\
PolyEnc ($S$) & 27.1 & 36.1 & 30.9\small$\pm$0.3 & 29.8 & 37.1 & 33.0\small$\pm$0.2 & 69.7& 51.7 & 59.3\small$\pm$0.2 & 73.8\small$\pm$0.2 \\
PolyEnc ($S/P/D$) & 37.7 & 31.0 & 34.0 \small$\pm$0.7 & 40.7 & 31.2 & 35.3\small$\pm$0.4 & 78.0 & 49.1 & 60.3\small$\pm$0.2 & 66.0\small$\pm$0.2 \\
DCA ($S$) & 28.2 & 38.5 & 32.5\small$\pm$0.4 & 30.7 & 39.0 & 34.3\small$\pm$0.4 & 70.9 & 53.0 & 60.6\small$\pm$0.1 & 77.2\small$\pm$0.1 \\
DCA ($S/D$) & 35.7 & 39.3 & 37.4\small$\pm$0.6 & 37.7 & 38.6 & 38.1\small$\pm$0.4 & 78.2 & 51.8 & 62.3\small$\pm$0.1 & 75.6\small$\pm$0.1 \\
DCA ($S/P$) & 47.1 & 33.8 & 39.4\small$\pm$0.3 & 48.3 & 32.1 & 38.6\small$\pm$0.2 & 75.7 & 49.4 & 59.8\small$\pm$0.1 & 76.4\small$\pm$0.1 \\
DCA ($S/P/D$) & 45.4 & 33.2 & 38.3\small$\pm$0.1 & 46.9 & 31.9 & 38.0\small$\pm$0.2 & 80.4 & 49.5 & 61.3\small$\pm$0.4 & 74.4\small$\pm$0.6\\ 
DCA (\tiny{$S/P/D$, no filters}) & 40.2 & 31.8 & 35.5\small$\pm$0.9 & 42.7 & 31,4 & 26.2\small$\pm$0.5 & 80.0 & 50.4 & 61.8\small$\pm$0.5 & 74.4\small$\pm$0.6 \\
DCA ($P/D$) & 25.4 & 26.8 & 26.0 & 28.5 & 27.9 & 28.2 & 74.7 & 45.7 & 56.8 & 39.5 \\
\end{tabular}
\caption{Test set results on PARSEME 1.1 English and DiMSUM for MWE identification, and the English all-words WSD task. For MWE identification, all Bi-encoder (BiEnc) and and Poly-encoders (PolyEnc/DCA) function as a final filter in the rule-based pipeline. Letters after system entries indicate training data, where $S$ = SemCor, $P$ = PARSEME and $D$ = DiMSUM. For example, ($S/P/D$) means trained on SemCor and finetuned on PARSEME and DiMSUM. Scores marked with the asterisk $^{*}$ come from pipeline configurations that did not produce valid output according to the DiMSUM scorer and had to be approximated, so they may be off by 1--2 F1 points. }
\label{tab:appendix}
\end{table*}}



\end{document}